\documentclass[3p,times]{elsarticle}

\usepackage{ecrc}

\volume{00}

\firstpage{1}

\journalname{Intelligent Medicine}

\runauth{}

\jid{procs}

\jnltitlelogo{IM}

\CopyrightLine{2022}{Published by Elsevier Ltd.}

\usepackage{amssymb}

\usepackage[figuresright]{rotating}
\usepackage{amsmath}
\usepackage{graphicx}
\usepackage{hyperref}
\usepackage{todonotes}
\usepackage{color} 
\usepackage{math}
\usepackage{float}
\usepackage{subfigure}

\usepackage{booktabs}

\usepackage{algorithmicx}

\makeatletter  
\newif\if@restonecol  
\makeatother

\usepackage[linesnumbered,ruled,vlined]{algorithm2e}
\usepackage{algpseudocode}  
\usepackage{amsmath}  

\usepackage{multirow}

\usepackage{lineno}

\begin{document}

\begin{frontmatter}



\dochead{}

\title{Application of Transfer Learning and Ensemble Learning in Image-level 
Classification for Breast Histopathology}


\author[prgA]{Yuchao Zheng}
\author[prgA]{Chen Li}\cormark[1]\ead{lichen201096@hotmail.com}
\author[prgA]{Xiaomin Zhou}
\author[prgA]{Haoyuan Chen}
\author[prgA]{Hao Xu}
\author[prgA]{Yixin Li}
\author[prgA]{Haiqing Zhang}
\author[prgB]{Xiaoyan Li}
\author[prgB]{Hongzan Sun}
\author[prgC]{Xinyu Huang}
\author[prgC]{Marcin Grzegorzek}

\address[prgA]{Microscopic Image and Medical Image Analysis Group, 
College of Medicine and Biological Information Engineering, \\ 
Northeastern University, Shenyang, China}
\address[prgB]{China Medical University, Shenyang, China}
\address[prgC]{Institute of Medical Informatics, University of Luebeck, Luebeck, Germany}

\begin{abstract}

\textbf{Background:} Breast cancer has the highest prevalence in women globally. The classification and 
diagnosis of breast cancer and its histopathological images have always been a hot 
spot of clinical concern. In Computer-Aided Diagnosis (CAD), traditional classification 
models mostly use a single network to extract features, which has significant limitations. 
On the other hand, many networks are trained and optimized on patient-level datasets, 
ignoring the application of lower-level data labels. 

\textbf{Method:} This paper proposes a deep ensemble 
model based on image-level labels for the binary classification of benign and malignant lesions of breast histopathological images. First, the BreaKHis dataset is randomly divided into 
a training, validation and test set. Then, data augmentation techniques 
are used to balance the number of benign and malignant samples. Thirdly, considering 
the performance of transfer learning and the complementarity 
between each network, VGG16, Xception, ResNet50, DenseNet201 are selected as the 
base classifiers.

\textbf{Result:} In the ensemble network model with accuracy as the weight, 
the image-level binary classification achieves an accuracy of $98.90\%$. In order to 
verify the capabilities of our method, the latest Transformer and Multilayer Perception 
(MLP) models have been experimentally compared on the same dataset. Our model wins with 
a $5\%-20\%$ advantage, emphasizing the ensemble model's far-reaching significance 
in classification tasks.

\textbf{Conclusion:} This research focuses on improving the model's classification performance with an ensemble algorithm. Transfer learning plays an essential role in small datasets, improving training speed and accuracy. Our model has outperformed many existing approaches in accuracy, providing a method for the field of auxiliary medical diagnosis.

\end{abstract}


\end{frontmatter}




\section{Introduction}

\par{Breast cancer is one of the most common malignant epithelial tumors in the world. According 
to the latest statistics from the International Agency for Research on Cancer (IARC) of 
the World Health Organization (WHO), there are 2.26 million new cases of breast cancer 
worldwide, surpassing the 2.2 million cases of lung cancer. At the end of 2020, breast 
cancer officially replaced lung cancer and became the world's largest 
cancer~\cite{2020Breast}. Early treatment of breast cancer is vital, and doctors need to choose an 
effective treatment plan based on its malignancy. Therefore, the detection of breast 
cancer, the distinction between cancerous structure and the 
identification of its malignant degree is valuable. Previously, there were 
many techniques for detecting breast cancer, including Magnetic Resonance Imaging 
(MRI)~\cite{murtaza2020deep}, Computed Tomography (CT), Positron Emission Tomography 
(PET)~\cite{domingues2020using}, Ultrasound technology (US)~\cite{kozegar2020computer}, 
Mammograms (X-ray)~\cite{moghbel2020review} and Breast Temperature 
Measurement~\cite{moghbel2013review}, etc. 
\par{At present, histopathological diagnosis is 
generally regarded as a ``gold standard''~\cite{de2019histopathologic}. Pathologists 
need to observe the tissue lesions under the microscope to determine the cancerous area and the 
degree of malignancy based on tissue structure, the nucleus and cytoplasm, and the growth pattern of the cells. To better analyze the different components of the tissue under the microscope, histopathologists usually stain the cut tissue.
In all staining methods, Hematoxylin and Eosin staining has been frequently used for more than a 
hundred years, abbreviated as H\&E~\cite{gurcan2009histopathological}. 
Fig.~\ref{FIG:BreastTissue} shows images of different types of breast tissues stained 
with H\&E. Blue is the color of cell nuclei in excised tissues stained with hematoxylin, 
and pink is the color of other structures (cytoplasm, matrix, etc.) stained with eosin. According to their biological behaviors, breast lesions can be divided into benign lesions (Fig.~\ref{FIG:BreastTissue} (b)) and malignant lesions (Fig.~\ref{FIG:BreastTissue} (c)(d)). Benign lesions generally grow slowly and do not exhibit an invasive growth pattern, while malignant tumor tends to metastasize to lymph nodes and distant organs.}

\begin{figure}[!h] 
	\centering 
	\includegraphics[width=1\textwidth]{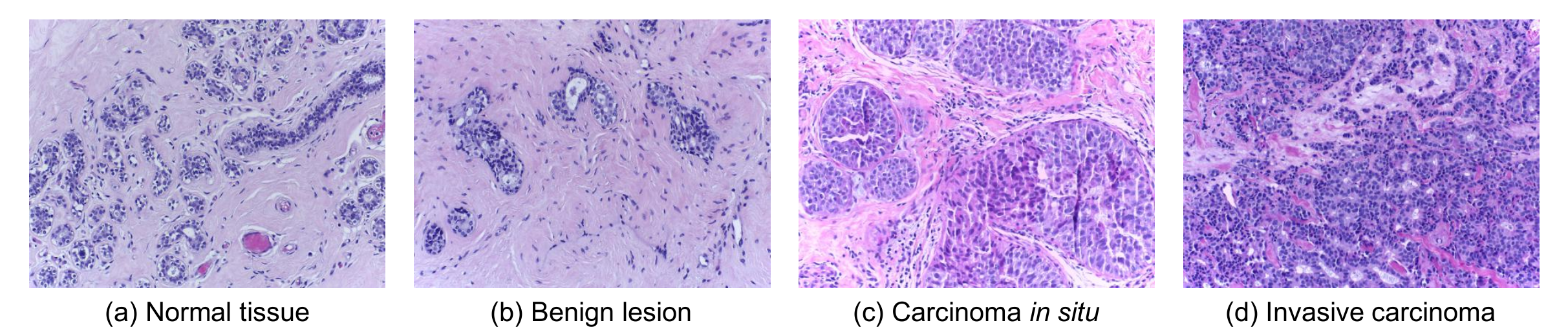}
	\caption{Breast tissue types based on H\&E staining.} 
	\label{FIG:BreastTissue}
\end{figure}

\par{However, it is tough for a histopathologist to observe the tissue with 
the naked eyes and manually analyze the visual information based on prior medical 
knowledge. For one thing, manual analysis takes much time. The histopathological 
image itself will lead to complexity and diversity due to subtle differences, cell overlap, 
uneven color distribution and others. For another, the objectivity of this kind 
of analysis is unstable. The reason is that it depends largely on the experience, 
workload, and emotion of the histopathologist~\cite{li2019survey}. Authoritative 
pathologists still need highly professional training and rich experience to make a 
reliable diagnosis of patients, but the expertise and experience are also quite hard to inherit or innovate. Therefore, there is an urgent need for a system that can 
realize \emph{Breast Histopathology Image Classification} (BHIC) to distinguish between 
cancerous tissue (malignant tissue) and non-cancerous tissue (benign tissue) to help 
pathologists make the diagnosis process more efficient and straightforward.}

\par{With the enhancement of computer calculating capacity and the continuous improvement 
of storage performance, CAD technology has quickly become an indispensable technology in 
the medical field, especially in the realm of histopathological image analysis. Currently, 
the medical reliance on CAD systems has increased year by year. The histopathological 
images can be quickly filtered and pre-classified by the CAD system, and then clinical 
doctors can obtain the second idea of early diagnosis. This helps reduce the burden 
on pathologists and improves the efficiency of work. Meanwhile, it can also concentrate the 
influence of pathologists on subjectivity and personal experience differences. With the 
assistance of application technology in the CAD system, the difference between observers has 
been further avoided. Above all, It is a work with research significance and broad 
application prospects.}

\par{Faced with this situation, much research around the world is ongoing. However, most 
of the existing methods are based on a single classifier. On the other hand, very few classification models based on image-level can achieve good classification results. This is because image-level classification is more complex: 
Images under the same patient-level label are related, making it easier to obtain better 
classification results; in contrast, image-level images only have relatively independent 
labels, and there is no correlation between each label and less practical information 
than patient-level images. While patient-level images are of great value in medical diagnosis, they sometimes add to the problem. For example, in the Medical Image Retrieval System~\cite{ghosh2011Review}, when the doctor encounters a difficult disease, a query image can be sent to the system for assisted diagnosis through the retrieval system. It is very inconvenient to upload full slices or a whole set of data, and image-level data can greatly simplify upload, so image-level analysis is also very valuable. Therefore, this paper proposes an image-level classification 
method based on transfer learning and ensemble learning. It is believed that based on 
the original research, ensemble learning and transfer learning can provide users with 
easy-to-operate software and achieve better classification results. Since transfer 
learning utilizes a pre-trained model, it can significantly adapt to a small dataset 
in a short time~\cite{Rahaman-2020-IOCS}. Ensemble learning can complement the advantages 
of multiple networks, 
significantly improving the accuracy and generalization ability of the model. At the 
same time, compared with the classification task under the patient-level label, the 
classification of the lower image-level is more challenging. The workflow is shown in 
Fig.~\ref{FIG:WorkflowWork}.} 
\begin{figure}[!h] 
	\centering 
	\includegraphics[width=0.98\textwidth]{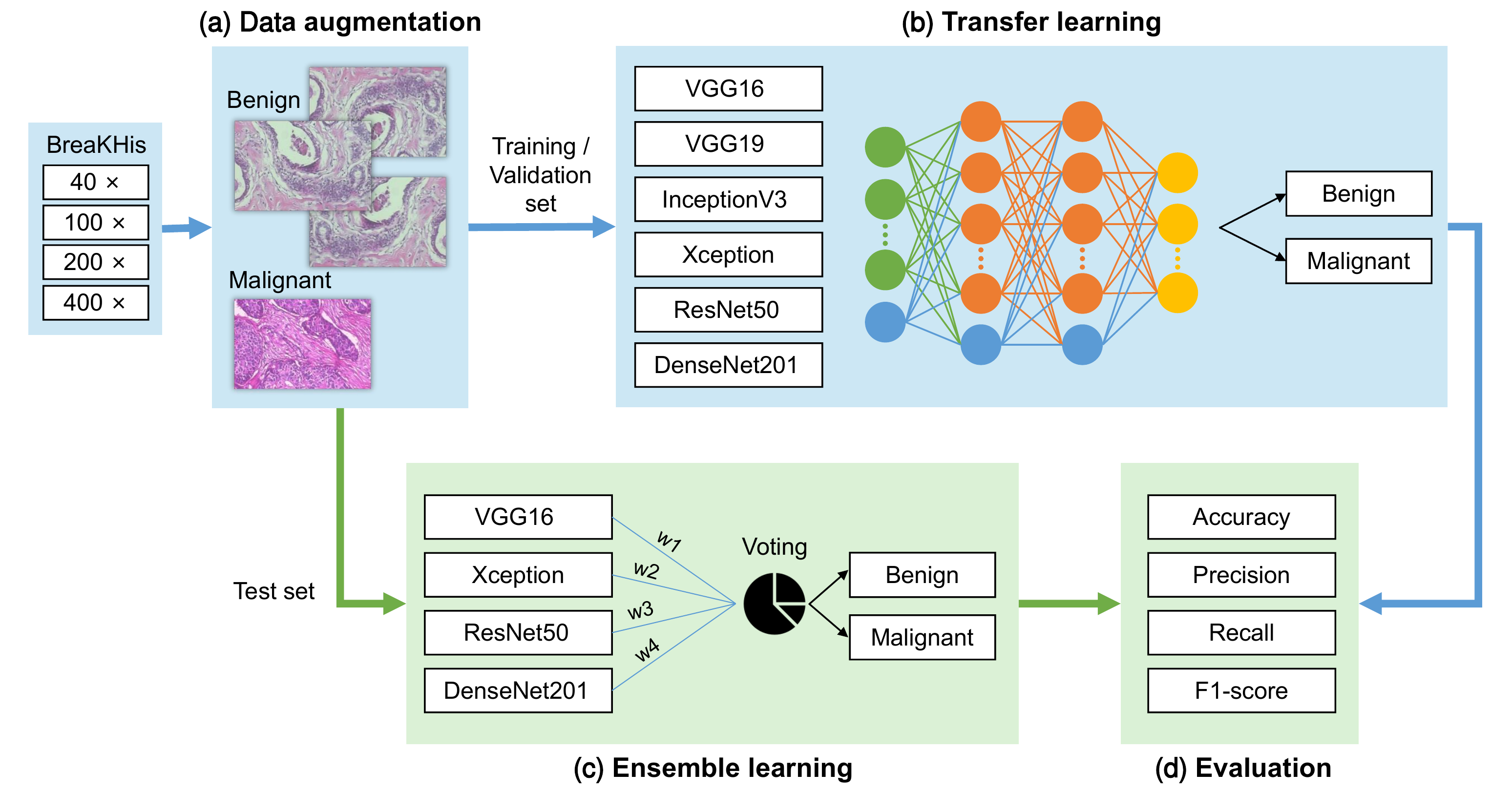}
	\caption{Workflow of the proposed work.} 
	\label{FIG:WorkflowWork} 
\end{figure}

As the workflow presented, the basic framework of this method is mainly composed 
of four parts:

(a) Data augmentation. All of the images from the BreaKHis dataset are divided into training, validation and test sets at the ratio of 7:1:2. Then the benign images are augmented through mirror flipping in order to balance the benign and malignant sample quantities.

(b) Transfer learning. The convolutional neural networks (CNNs), including VGG16, VGG19, ResNet50, InceptionV3, Xception and DenseNet201 are used for transfer learning to get applicable networks. According to the evaluation indicators, four out of six models are chosen as the base classifiers for the next process.

(c) Ensemble learning. This strategy based on the weighted voting method is used to further improve the classification performance. And the accuracy, one of the evaluation indicators, is chosen as the weight.

(d) Evaluation. The accuracy, precision, recall and F1-score are used as indicators to measure the classification ability of the whole algorithm.

\par{The main contributions of this study are summarized as follows:} 
\par{(1). This subject has technical advantages. A new framework is proposed to solve 
the classification problem of breast cancer histopathological images. In our previous 
work~\cite{zhou2020comprehensive}, the application of classic neural networks and deep 
learning in breast tissue pathological images from 2012 to early 2020 has been summarized 
in detail. At the same time, combined with the analysis of the literature in recent 
years, it is found that there are not many ideas for adopting the method of ensemble 
learning, which is very promising for research.}

\par{(2). Sufficient experiments are guaranteed for results. In order to prevent the 
singularity and limitation of a single classifier, six CNNs are used for training. Then, 
the best four neural networks are ensembled based on the weighted voting method. Through 
multiple experiments, we find that when the accuracy is used as the weight, the classification 
result is the best.}

\par{(3). Excellent classification accuracy is achieved. The classification system 
can effectively overcome the problem of small datasets, while significantly reducing training 
time. On this basis, a classification accuracy of $98.90\%$ is obtained.}

\par{This paper is divided into the following chapters: Section~\ref{section:2} 
introduces the related technologies of breast histopathological image classification. 
Section~\ref{section:3} discusses the proposed image classification method based on 
transfer learning and ensemble learning in detail. Section~\ref{section:4} introduces 
the experimental settings, process, analysis and limitation of the results. 
Section~\ref{section:5} summarizes this paper and puts forward the future plan under 
this topic.}

\section{Method}
\label{section:2}

\subsection{Related work}

\par{This part mainly introduces the related works involved in the paper and the research status in recent years.}

\subsubsection{Breast cancer and the application of CAD in breast histopathology}

\par{Among women, breast cancer accounts for one-quarter of cancer cases and one-sixth 
of cancer deaths. It ranks first in most countries (159 out of 185 countries) in morbidity 
and first in mortality among 110 countries~\cite{sung2021global}. In fact, women in all 
regions of the world are at risk of breast cancer at any age after puberty, and the 
incidence will increase in their later years.}

\par{The onset of breast cancer is caused by negative mutations in certain genes, 
after which breast tissue cancer is evolved into malignant tumors. There are two major 
types of breast cancer in medicine: carcinoma \emph{in situ} and invasive 
carcinoma. Carcinoma \emph{in situ} refers to the early stage of breast cancer where 
the growth of cancer cells is confined to ducts or lobules. Usually, the symptoms do 
not manifest themselves, and the possibility of spreading or metastasis is small. 
Over time, these cancer cells \emph{in situ} may gradually develop and invade the 
surrounding breast tissue and then spread to nearby lymph nodes (specific regional 
metastasis) or other organs in the body (distant metastasis). This is called invasive 
carcinoma.}

\par{There are many pathogenic factors of breast cancer, mostly related to gender and age. 
Other reasons mainly include obesity~\cite{renehan2008body}, lack of healthy 
exercise~\cite{mctiernan2003recreational}, a high-protein diet, such as eating red meat 
with exogenous hormones or carcinogenic by-products~\cite{levine2014low}, 
alcoholism~\cite{collaborative2002alcohol}, smoking~\cite{us2014health}, and using 
oral contraceptives~\cite{hunter2010oral}. These risk factors can be intervened 
through health education in clinical practice and public health programs. Unfortunately, 
even if all potential variables can be controlled, the risk of breast cancer can only 
be reduced by up to $30\%$.}

\par{Numerous significant advances in computer science and medicine have created exciting 
opportunities for medical experts and intelligent systems. CAD is the most direct and 
obvious result of the organic integration of the two. CAD does not rely on the analysis 
and skills of individual healthcare professionals and can make more objective and 
fasterdiagnostic decisions. In addition, CAD can narrow the gap between experienced and 
inexperienced medical personnel in diagnosis.}

\subsubsection{Deep learning model}

\par{Since 2012, the application of deep neural network (DNN) technology has become more 
and more extensive, including image preprocessing, feature extraction, image postprocessing, 
and classifier design. This development trend can be attributed to the rapid improvement 
of hardware performance, which provides high feasibility for realizing DNN algorithms 
with high computational complexity. In addition, an automatic feature extraction method 
can be realized by DNN, through which it can be more robust to extract complex microscopic 
breast tissue morphological features and structures. In the era of DNN, CNN is often preferred in image classification. Because of its superior performance in related fields, such as face recognition~\cite{co2017Face}, autonomous driving~\cite{gao2018Object}, COVID-19 image analysis~\cite{maram2021CovidXrayNet} and so on, CNN is also increasingly preferred in BHIC.}

\par{CNN is first proposed in~\cite{lecun1998gradient}, 
in which the backpropagation algorithm to the training of neural network structure is 
applied. Compared with traditional image classification methods, CNN-based deep learning 
can automatically learn features from massive amounts of data. Moreover, this method 
effectively reduces the interference of subjective factors in the traditional method. 
At the same time, CNN has been well applied in many fields, including object or human 
body recognition, target tracking and detection, image segmentation and classification, 
natural language processing, etc. It has laid a good foundation for the application 
of CNN in BHIC.}

\par{VGGNet~\cite{simonyan2014very} is one of the top two networks in the ImageNet Large-Scale 
Visual Recognition Challenge (ILSVRC)~\cite{russakovsky2015imagenet} in 2014. The VGG16 and 
VGG19 models are simple in structure, but consume many resources, training time, and 
storage capacity. The Inception module is the core component of 
GoogLeNet~\cite{szegedy2015going}. Based on the first two versions, InceptionV3 uses the 
idea of factorization to split a two-dimensional convolution into two smaller modules: 
$n \times n$ convolution into $1 \times n$ and $n \times 1$ convolution. It is beneficial 
to reduce the number of parameters, speed up the calculation and further increase the 
depth and nonlinearity of the network. The Xception network is an improved version of 
InceptionV3 and is also regarded as an ``extreme'' version of the 
Inception~\cite{chollet2017xception}. The network introduces a deep separable convolution 
model, and an excellent effect can be obtained after comprehensive training data. One of 
the basic modules that make up ResNet~\cite{he2016deep} is called the residual block, 
which can effectively solve the problem of gradient disappearance in deep networks. 
DenseNet~\cite{huang2017densely} is well-known for its particular structure. In forward 
propagation, each layer is directly connected with all previous layers, and the input 
of each layer comes from the output of all previous layers. Compared with the deep 
residual network, the number of parameters that need to be trained is much lower than 
that of ResNet to achieve the same accuracy.}

\par{With the proliferation of deep learning, people are increasingly pursuing stable models with good performance in all aspects, but the actual results are not so ideal. Therefore, the ensemble approach emerges as The Times require, which can complement every single classifier well. The work of \cite{Cao2020Ensemble} summarizes the recent applications in bioinformatics based on ensemble deep learning. Paper \cite{Yongquan2021Discussion} further discusses the development and limitations of ensemble learning in the era of deep learning, which is instructive for our research.}

\subsubsection{The development of breast histopathological image analysis}

\par{In the past ten years, CNN has had an outstanding performance in BHIC. In \cite{petushi2006large}, a third-party software (LNKNet package) containing a neural 
network classifier is applied to evaluate two specific textures, namely the quantity 
density of the two landmark substances, and a $90\%$ classification accuracy of 
breast histopathology images is achieved.}

\par{In the paper~\cite{singh2010breast}, to classify breast histopathological images 
stained by H\&E into four types, eight features and a three-layer forward/backward 
artificial neural network (ANN) classifier are applied. Finally, a classification 
accuracy of about 95$\%$ is obtained.}

\par{In~\cite{zhang2011breast,zhang2013breast1,zhang2013breast2}, an automatic 
classification scheme is given. ``Random subspace ensemble'' is used to select and 
aggregate the designed classifiers. Moreover, a classification accuracy of 95.22$\%$ 
is obtained on a public image dataset.}

\par{In the paper~\cite{shukla2017classification}, morphological features are extracted 
to classify cancer cells and lung cancer cells in histopathological images. In the 
experiment, the multi-layer perceptron based on the feedforward ANN model obtains 
$80\%$ accuracy, 82.9$\%$ sensitivity, and 89.2$\%$ AUC successfully.}

\par{In~\cite{das2017classifying}, a CNN-based classifier of full-slice histopathological 
images is designed. First, the posterior estimate is obtained, which is derived from 
the CNN with specific magnification. Then, the posterior estimates of these random 
multi-views are vote-filtered to provide a slice-level diagnosis. Finally, 5-fold 
cross-validation at the patient-level is used in the experiment, with an average 
accuracy of 94.67$\%$, a sensitivity of 96$\%$, a specificity of 92$\%$, and an 
F1-score of 96.24$\%$.}

\par{In the paper~\cite{gour2020residual}, a ResHist model is designed to classify benign 
and malignant breast histopathological images, which is based on a 152-layer CNN with 
residual learning. In the experiment, the histopathology images are first enhanced, 
and then the ResHist model is trained end-to-end in the enhanced dataset by supervised 
learning. After testing by the ResHist model, the best accuracy of 92.52$\%$ and the F1-score 
of 93.45$\%$ are achieved. In addition, to study the recognition ability of the ResHist 
model for in-depth features, researchers input the extracted feature vectors into KNN, 
Random Forest~\cite{breiman2001random}, secondary discriminant analysis, and SVM 
classifiers. The best accuracy of 92.46$\%$ can be obtained when the deep functions 
are fed back to the SVM classifier.}

\par{In~\cite{yan2020breast}, a new hybrid convolutional and cyclic DNN is created to 
classify breast cancer histopathological images. The paper uses fine-tuned InceptionV3 
to extract features for each image block. Then, the feature vectors are input to the 
4-layer bidirectional long and short-term memory network~\cite{schuster1997bidirectional} 
for feature fusion. In the experiment, the image is completely classified, and the 
average accuracy is 91.3$\%$. It is worth noting that in this paper, a new dataset 
containing 3,771 histopathological images of breast cancer is published, which covers 
as many different sub-categories of different ages as possible.}

\par{In~\cite{kausar2019hwdcnn}, the deep CNN model based on the Haar wavelet is introduced 
to classify breast tissue pathological images, greatly reducing the calculation time 
and improving the classification accuracy. In the experiment, the accuracy of 98.2$\%$ 
and 96.85$\%$ are achieved for the BACH dataset of four and two types and the BreaKHis 
dataset of multiple types, respectively.}

\par{In the classification work of~\cite{li2019classification}, an analysis technique 
based on deep learning is proposed. The technique first uses a method based on CNN 
and $k$-means for screening. Then ResNet50 is used to extract features, and P-norm 
pooling is used to get the final image features. In the end, SVM is applied for the 
final image classification, with an accuracy of 95$\%$.}

\par{In~\cite{wang2018classification}, to distinguish four types of breast cancer in 
histopathological images, a novel deep learning method is introduced. In the method, 
hierarchical loss and global pooling are applied. VGG16 and VGG19 models are used as 
the base deep learning network, and a dataset containing 400 images is used for testing. 
Finally, an average accuracy of around 92$\%$ is obtained.}

\par{At present, there are a few types of research on the classification of breast 
histopathological images using ensemble learning methods. Among them, the classification 
based on patient-level labels is commonly studied, such as research in 
Table.~\ref{TABLE:PatientLevelClassification}.} 

\begin{table}[!h]
\centering
\caption{Research on the patient-level classification of breast histopathological images.}
\begin{tabular}{llllr}
\toprule 
Year & Method                                            & Dataset              & Transfer Learning CNNs               & Accuracy \\ \midrule 
2019 & Ensemble of DNNs \cite{kassani2019classification} & BreaKHis             & VGG-19, MobileNet-V2, DenseNet-201   & 98.13$\%$   \\
2019 & EMS-Net \cite{YANG2019EMSNet}                     & BreaKHis(40 ×)    & ResNet-101, ResNet-152, DenseNet-161 & 99.75$\%$   \\
2020 & Ensemble of DNNs \cite{Anda2020His}               & Collected 544 Images & VGG-16, VGG-19                       & 95.29$\%$   \\
2021 & MCUa \cite{Senousy2021MCUa}                        & BreaKHis(40 ×)    & ResNet-152,  DenseNet-161            & 100.00$\%$  \\
2021 & 3E-Net \cite{Senousy2021ENet}                    & BreaKHis(40 ×)    & DenseNet-161, 6 image-wise CNNs      & 99.95$\%$   \\ \bottomrule 
\end{tabular}
\label{TABLE:PatientLevelClassification} 
\end{table}

\par{From Table.~\ref{TABLE:PatientLevelClassification}, it can be seen that the existing 
methods have achieved outstanding results in patient-level classification tasks. Although 
patient-level classification is the most common clinical requirement, many studies 
classify breast histopathological images at the image level~\cite{Zhu-2021-HMEHE}. These 
tasks are generally combined with patient-level classification to train and test the 
proposed model, such as~\cite{Song2018Imagelevel,Li2020Imagelevel,hu2021Imagelevel}. 
Nevertheless, since the image-level classification is more complex than the patient-level, 
the classification accuracy is not relatively high in previous research. Therefore, this 
paper focuses on the challenge of image-level classification. At the same time, the 
characteristics of each transfer learning model are comprehensively considered, and 
ensemble learning is carried out based on their complementarity.}

\subsection{Our method}
\subsubsection{Transfer learning}

\par{Transfer learning is a method of applying knowledge or patterns learned in a 
specific field or task to different fields or problems~\cite{ribani2019survey}. 
Traditionally, machine learning algorithms have strict training and test data 
requirements, which require them to obey the same distribution. However, in reality, 
this hypothesis cannot be fully established, resulting in many restrictions on the 
practical application of machine learning. At the same time, the acquisition of 
sufficient training data is another big problem for researchers. Therefore, scholars 
have developed a special method to use a large amount of easy-collected data from 
different fields to train learners and apply them to other similar scenarios. This 
method is called transfer learning. In other words, if $X_{S}$ stands for the feature space and $X_{T}$ stands for the label space}, given a source domain 
$D_{S}=\{ X_{S}, f_{S}(x)\} $ and learning task $T_{S}$, a target domain 
$D_{T}=\{ X_{T}, f_{T}(x)\} $ and learning task $T_{T}$, transfer learning aims to 
help improve the learning of the target predictive function $f_{T}(x)$ in $D_{T}$ 
using the knowledge in $D_{S}$ and $T_{S}$, where $D_{S}\neq D_{T}$, or $T_{S}\neq T_{T}$. 
As for neural networks, researchers can directly use pre-trained models through a 
large number of ready-made datasets. Then, reusable layers are selected, and the 
output of these layers is used as input to train a network with fewer parameters 
and a smaller scale. This small-scale network only needs to clarify the internal 
relationships of specific problems and learn the implicit patterns through pre-trained 
models~\cite{shoeleh2017graph}.

\par{At present, there are two main methods of applying transfer learning: one is to 
fine-tune the parameters of the pre-trained network according to the required task; 
the other is to use the pre-trained network as a feature extractor and then use these 
features to train a new classifier. In this paper, we choose the former one.}

\par{The reasons why transfer learning is selected in this paper are as follows:} 
\par{(1). Well-labeled histopathological images of breast cancer are relatively limited. 
Due to the complexity of such images, it is costly for medical experts to annotate data. 
Therefore, few publicly large-scale image datasets are available. Fortunately, transfer 
learning can effectively overcome the problem of insufficient 
data~\cite{hadad2017classification}.}

\par{(2). In the classification task of histopathology images, most of the preprocessing 
models come from ILSVRC. Because of their stability in specific challenges, they can 
be safely applied in breast cancer classification tasks.}

\par{(3). Transfer learning helps to improve accuracy or reduce the training 
time~\cite{sarkar2018hands}, which is a crucial reason why transfer learning is 
widely welcomed.}

\par{In this paper, the VGG series, Inception series, ResNet series, and DenseNet series 
are carefully studied and compared, and six CNNs are selected to classify 
breast histopathological images into benign and malignant tumors. They are: VGG16, 
VGG19, InceptionV3, Xception, ResNet50 and DenseNet201. These classic networks are 
selected because, on the one hand, these networks have passed numerous classification 
tasks and have shown high accuracy and stability in various datasets. In addition, the 
advantages and shortcomings of these networks are systematically considered, and it 
is highly possible to build a comprehensive ensemble network by using their complementarity.}

\par{For VGGNet, the VGG16 and the VGG19 network model are selected, because of their 
simple network structure and excellent learning performance on many tasks. In terms of the classiﬁcation ability and parameter quantity, InceptionV3 and Xception network models are adopted for the Inception series. In the ResNet series and the DenseNet series, considering the calculation ability and other issues, we finally choose the ResNet50 
and the DenseNet201 network model to conduct the research. In summary, six 
CNN models of VGG16, VGG19, InceptionV3, Xception, ResNet50, and DenseNet201 are 
applied to the classification of breast histopathology images.}

\par{In the transfer learning based on these six CNNs, first of all, the massive labeled 
images in the ImageNet dataset are applied separately, so these networks have good 
classification capabilities. Then, the front ends of the trained model parameters in 
these six CNNs are all frozen, so as not to destroy any information they contain in 
future training. After that, the augmented breast histopathological images and their 
labels are used to fine-tune the fully connected layers at the back end of the six 
networks. Finally, the classification ability learned on ImageNet can be transferred 
to the given pathological slices.}

\subsubsection{Ensemble learning}

\par{In the past few decades, ensemble learning has received more and more attention 
in the field of computational intelligence and machine learning. Ensemble learning is 
a process in which multiple models, such as classifiers, are combined according to a 
certain method to solve a specific intelligent computing problem. It is mainly used 
to improve the performance of the final model, such as classification, prediction, 
and function estimation, or to reduce the impact of improper selection of the basic model.}

\par{The general framework of ensemble learning is summarized as follows: First, a 
group of individual learners is generated; after that, the learners are effectively 
combined through a specific strategy; finally, the expected experimental results can 
be obtained~\cite{2009Ensemble}. Individual learners are usually generated from training 
data by existing learning algorithms (such as decision tree and error propagation neural 
network). Among them, using individual learners of the identical type (such as all 
neural networks) is called homogeneous ensemble learning, and applying various 
individual learners is called heterogeneous ensemble learning. In this paper, 
homogeneous ensemble learning is used.}

\par{Recent fast ensemble deep learning techniques~\cite{Yang2020FTBME, Yang2021Local} can be applied in whole slide medical image analysis (f.e. predicting pCR from H\&E stained whole slide images~\cite{Li2021Deep}) to reduce time and space overheads at the expense of certain accuracy. However, we believe that performance improvement is more critical in this task, so we choose some usual ensemble methods. There are three general combination strategies: voting, 
averaging, and stacking. Nevertheless, for the task of this paper that aims to 
achieve the binary classification of breast histopathological images, the voting 
method is the most commonly used and easily manipulated one. This method refers to 
the assumption that there are $T$ different classifiers ${h_{1}, h_{2},..., h_{T}}$, 
and our goal is to predict the final category from the $l$ category markers 
${c_{1}, c_{2},..., c_{l}}$ based on the output of the classifier. Typically, for 
the sample $x$, the output result of the classifier $h_{i}$ is an $l$-dimensional 
label vector $(h_{i}^{1}(x), h_{i}^{2}(x),..., h_{i}^{l}(x))^{T}$, where 
$h_{i}^{j}(x)$ is the prediction output of the $h_{i}$ classifier on the class 
$c_{j}$ label.}

\par{Voting can be divided into three types. Absolute majority voting: Different 
categories will be marked by each classifier for voting. If the final number of 
votes obtained by a certain type of mark exceeds $\frac{1}{2}$, the category is 
regarded as the final output result; if the votes for all the categories do not 
exceed $\frac{1}{2}$, the forecast is rejected. Relative majority voting: The side 
with the most votes is deemed the winner; when there is a tie, one side is chosen 
arbitrarily. Weighted voting: Different from the previous two, the corresponding 
weights are assigned to different classifiers; if the classifier performs better, 
a higher weight is assigned. Finally, the weighted votes of each category are summed, 
and the one corresponding to the maximum value is regarded as the final result.}

\par{Although there are endless ensemble learning methods, considering their complexity 
and ease of operation, the basic weighted voting strategy is applied in this paper, 
and the formula is Eq.~(1). When given appropriate weights, weighted voting can be 
both superior to the individual classifier with the best classification result, and 
at the same time, superior to the absolute majority voting.} 
\begin{equation}
H(x)=c_{arg_{j}max}\sum_{i=1}^{T}w_{i}h_{i}^{j}(x)
\label{Eq:1}
\end{equation}

\par{Among them, $w_{i}$ represents the weight of the classifier $h_{i}$. In practical 
applications, similar to the weighted average method, the weight coefficients are 
often normalized and are constrained to be $w_{i}\geq 0$ and $\sum_{i=1}^{T}w_{i}=1$. 
Getting the right weight is very important. Suppose $l=(l_{1}, l_{2},...,l_{T})^{T}$ 
is the output of the individual classifier, where $l_{i}$ represents the prediction 
result of the class label of the classifier $h_{i}$ on sample $x$. Let $p_{i}$ be 
the precision of $h_{i}$, the combined output of the category label $c_{j}$ can be 
expressed as Eq.~(2) using a Bayesian optimal discriminant function.} 
\begin{equation}
H^{j}(x)=\log (P(c_{j})P(l|c_{j}))
\label{Eq:2}
\end{equation}

\par{When it is assumed that the output conditions of individual classifiers are 
independent, Eq.~(2) can be reduced to Eq.~(3).} 
\begin{equation}
H^{j}(x)=\log P(c_{j})+\sum_{i=1}^{T}h_{i}^{j}(x)\log \frac{p_{i}}{1-p_{i}}
\label{Eq:3}
\end{equation}

\par{When the first term of the above equation does not depend on the individual 
classifier, the optimal weight of the weighted voting can be obtained from the 
second term and satisfies Eq.~(4).}
\begin{equation}
w_{i}\propto\log \large\frac{p_{i}}{1-p_{i}} 
\label{Eq:4}
\end{equation}

\par{Therefore, the optimal weight needs to be consistent with the performance of 
the individual classifier, which is based on the case where the output of the 
individual classifier is independent of each other. However, in the actual work of 
this paper, the classifier is trained for the same problem. The output is usually 
strongly correlated, and the independence assumption is not valid. So the weight is 
set based on the evaluation indicators of the classifier, which is also a commonly 
used method of weighting.}

\section{Result}
\label{section:3}

\subsection{Experimental settings}
\subsubsection{Image dataset}

\par{To implement the proposed model, a practical open-source BreaKHis dataset is used 
in the research of this paper \cite{spanhol2015dataset}. BreaKHis consists of 7,909 
clinically representative microscopic images of breast tumor tissue, which are collected 
from 82 patients using four magnifications.}

\par{Data source: The sample comes from a biopsy section of breast tissue, which is 
marked by pathologists in Brazil's P\&D laboratory;}
\par{Staining method: H\&E staining;}
\par{Magnification: $50 \times$, $100 \times$, $200 \times$, $400 \times$;}
\par{Microscope: Olympus BX-50 system microscope, with 3.3 times magnification relay 
lens, connected with Samsung digital color camera SCC-131AN;}
\par{Camera pixel size: 6.5 $\mu m$;}
\par{Image size: $700 \times 460$ pixels;}
\par{Image format: *.png;}
\par{Pixel bit depth: 24 bits (RGB three-channel image, 8 bits per channel, 
$3 \times 8 = 24$);}

\par{Up to now, a total of 2,480 benign tumor samples and 5,429 malignant tumor 
samples have been collected in the BreaKHis dataset. Both of them are divided into 
different subtypes, respectively. As shown in Fig.~\ref{FIG:SubTypeSamples}, the 
first line is the four subtypes of benign tumors, including Adenosis (A), 
Fibroadenoma (F), Tubular Adenoma (TA), and Phyllodes Tumor (PT). The second line 
is the four subtypes of malignant tumors, including Ductal Carcinoma (DC), 
Lobular Carcinoma (LC), Mucinous Carcinoma (MC), and Papillary Carcinoma (PC).} 
\begin{figure}[!htbp] 
	\centering 
	\includegraphics[width=0.98\textwidth]{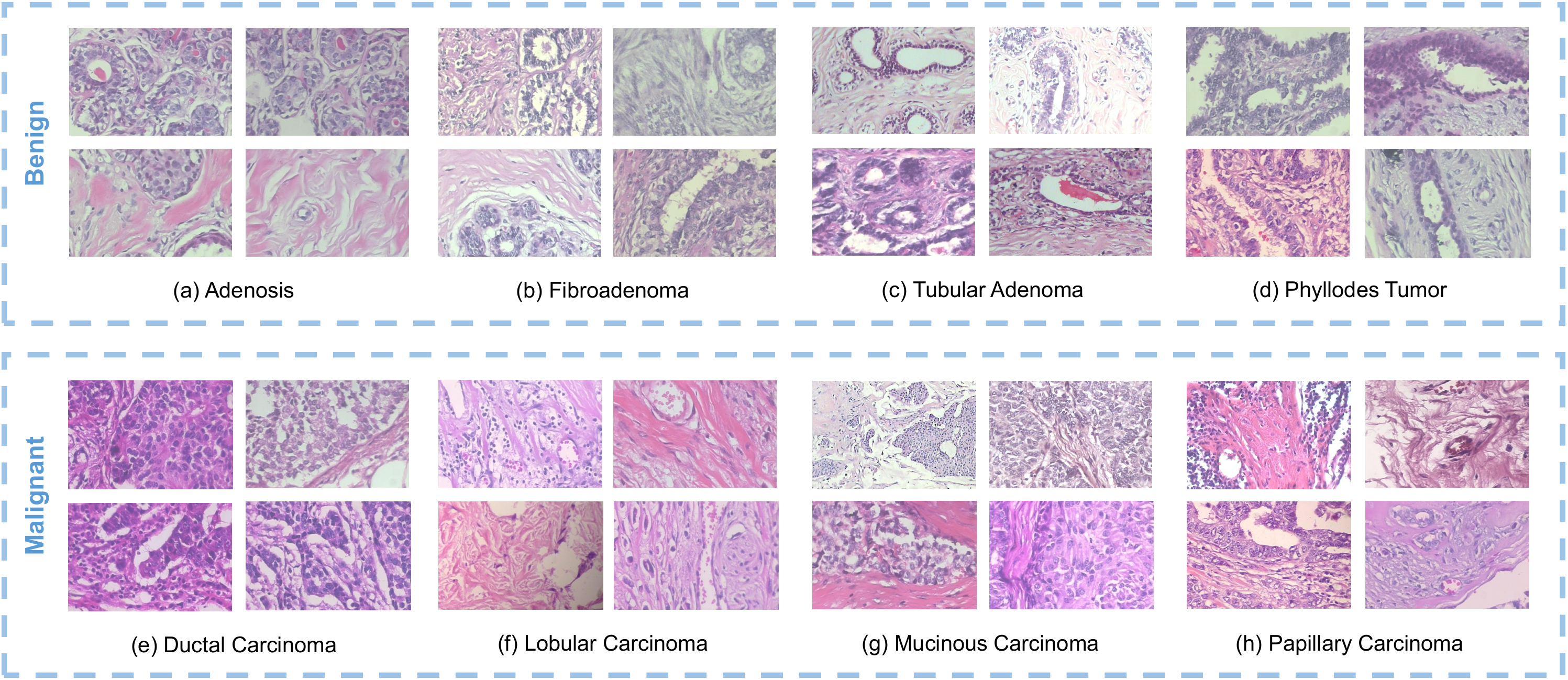}
	\caption{Benign/malignant tumor subtype samples} 
	\label{FIG:SubTypeSamples} 
\end{figure}

\par{Therefore, the dataset can be used for both binary classification and 
multi-classification tasks. In this paper, the binary classification of 
benign/malignant tumors under four magnifications is studied. Table \ref{TABLE:ImageDistribution} shows the distribution of samples in this dataset.} 
\begin{table}[!h]
\centering
\caption{Image distribution by magnification factor and class.}
	\begin{tabular}{llll}    
        \toprule 
		Magnification &   Benign \quad\quad\quad   & Malignant \quad\quad  & Total \quad \\    
        \midrule 
		$40 \times$               \quad\quad& 625    & 1,370   & 1,995   \\
		$100 \times$              \quad\quad& 644    & 1,437   & 2,081   \\ 
		$200 \times$              \quad\quad& 623    & 1,390   & 2,013   \\
        $400 \times$              \quad\quad& 588    & 1,232   & 1,820   \\
		Total               \quad\quad& 2,480  & 5,429   & 7,909   \\
        Number of patients  \quad\quad& 24     & 58      & 82      \\
		\bottomrule
	\end{tabular}  
\label{TABLE:ImageDistribution} 
\end{table}

\subsubsection{Setting of experimental data}

\par{In clinical work, patient-level classification is essential. In the early 
pre-experiment and review \cite{zhou2020comprehensive}, it has been found that many 
methods can achieve excellent classification results under the label of the patient. 
However, because the data information under the same patient label is highly 
correlated, the patient-level classification is not challenging enough. Therefore, 
the lower-level classification based on the image label is applied, and a better 
result is achieved in the preliminary experiment.}

\par{This experiment uses a cross-validation method to separate the BreaKHis dataset 
into three mutually exclusive subsets. The subsets are randomly selected from the 
dataset at a ratio of 7:1:2 and are separated into the training set, validation set 
and test set. However, according to the introduction of the dataset above, it is 
evident that the number of images under the two categories of benign (2,480) and 
malignant (5,429) is seriously imbalanced. Approximately $69\%$ of images are samples 
of malignant tumors. In order to tackle this problem, data augmentation strategies 
are applied to the benign tumor samples in all of the subsets, namely horizontal 
mirror flip and vertical mirror flip, with a total of 2,480 images generated. 
Afterward, 469 images are randomly selected from the data obtained through the 
vertical mirror flip through calculation. Finally, based on the original data 
volume (2,480), adding the horizontal mirror flip (2,480) and the selected vertical 
mirror flip dataset (469), the benign sample reaches 5,429 images. In this way, the 
number of images in both two categories is equal, with 10,858 images. The amount of 
samples in each subset is presented in Table~\ref{TABLE:Data}. In the early 
research~\cite{Li2020Generative}, we did data augmentation based on Generative Adversarial 
Network (GAN) and achieved relatively good results. However, the focus of this paper 
is not to generate a large number of new datasets, and only using mirror flipping 
can satisfy this experiment.} 
\begin{table}[!h]
\centering
\caption{Data setting.}
	\begin{tabular}{llll}   
		\toprule  
		\quad\quad\quad &   Benign \quad\quad\quad   & Malignant \quad\quad  & Total \quad \\    
		\midrule
		Training set    \quad\quad& 3,800   & 3,800   & 7,600    \\
		Validation set  \quad\quad& 543     & 543     & 1,086    \\ 
		Test set        \quad\quad& 1,086   & 1,086   & 2,172    \\
		\bottomrule 
	\end{tabular}  
\label{TABLE:Data} 
\end{table}

\subsubsection{Experimental environment}

\par{The experiments involved in this paper are all developed based on the Python 
language (version 3.6.8) under the Windows 10 operating system. All deep learning 
models involved in the experiment are developed (training, test) entirely by the 
Keras (version 2.24) framework. In addition, TensorFlow (version 1.12.0) is used 
as the backend of Keras. The specific conditions of the workstation parameters 
applied in the experiment are as follows: Intel(R) Core(TM) i7-8700 CPU (3.20 GHz), 
32GB RAM, NVIDIA GEFORCE RTX 2080 8 GB.}

\subsection{Evaluation metrics}

\par{In the course of our experiments, the performance of the classifier needs to be 
quantitatively evaluated. Appropriate evaluation indicators can reduce the deviation of 
algorithm differentiation. In this paper, confusion matrixes are used to analyze various 
evaluation measures. For the n-type classification problem, the confusion matrix is a 
table of size $n \times n$. An accurate classifier represents most samples along the 
diagonal of the matrix. In fact, the confusion matrix itself is not a performance 
indicator, but almost all frequently-used indicators are based on it. Table~\ref{tbl3} 
is a confusion matrix based on two classifications. Positive samples can be regarded 
as samples of research interest. Thus, we consider samples of benign tumors as positive 
and samples of malignant tumors as negative.} 
\begin{table}[!h]
\centering
\caption{Confusion matrix based on binary classification.}
	\begin{tabular}{lll}  
		\toprule    
		                                       &\quad\quad   Target class  &  \quad\quad    \\
        \multirow{-2}{*}{Output class}         &\quad\quad   Positive (P)        &\quad\quad  Negative (N)  \\
		\midrule  
		Positive (P)    &\quad\quad TP     & \quad\quad FP        \\
		Negative (N)    &\quad\quad FN     & \quad\quad TN        \\ 
		\bottomrule  
	\end{tabular}  
\label{tbl3} 
\end{table}

\par{TP (True Positive) is that an example is correctly identified as a positive; 
TN (True Negative) is that an example is correctly identified as a negative; 
FP (False Positive) is that negative cases are misclassified as positive ones; 
FN (False negative) is that positive instances are incorrectly classified as negative. 
This experiment evaluates the performance of six CNN classifiers mentioned above based 
on these four values.} 
\begin{table}[!h]
\centering
\caption{Evaluation metrics.}
	\begin{tabular}{ll}    
		\toprule   
		Assessment & \quad\quad\quad Formula  \\
		\midrule   
		\quad Accuracy         & \quad\quad\quad\large $\frac{TP+TN}{TP+FP+FN+TN}$                        \\ [6pt]
		\quad Precision        & \quad\quad\quad\large $\frac{TP}{TP+FP}$                                 \\ [6pt]
        \quad Recall           & \quad\quad\quad\large $\frac{TP}{TP+FN}$                                 \\ [6pt]
        \quad F1-score         & \quad\quad\quad\large $\frac{2\times P\times R}{P+R}$                    \\ [6pt]
		\bottomrule   
	\end{tabular}  
\label{TABLE:Evaluation} 
\end{table}

\par{For classification tasks, accuracy, precision, recall and F1-score are considered 
the most popular techniques for measuring the classification results. These evaluation 
indicators are described in Table~\ref{TABLE:Evaluation}. The accuracy is the ratio of 
the number of correctly classified samples to the total number of samples. The precision 
reflects the proportion of true positive cases in the cases that are judged to be 
positive, which represents how many benign tumors are predicted to be correct. The 
recall reflects the proportion of cases judged to be positive in the total number of 
positive cases. Low recall means that many patients with benign tumors are treated as 
cancer, which is a major medical incident. Therefore, recall is of great significance 
in medical image diagnosis. However, precision and recall are contradictory to a 
certain extent. If the precision is intended to be improved, it can be achieved by 
raising the criteria for positive classification. In other words, the number of samples 
predicted to be positive is reduced. Although the precision has improved, the recall 
will definitely decrease. On the contrary, if the recall needs to be improved, all 
samples can be marked as positive. In this way, the recall can reach $100\%$, but the 
precision will be absolutely reduced. Therefore, F$\beta$-score is introduced, which is a 
vital technique that balances precision and recall. When $\beta$ = 1, it is the 
harmonic average of the two values, namely the F1-score.}

\subsection{Deep learning algorithm}

\subsubsection{CNN training process}

\par{The training process of CNN is divided into two stages: forward propagation 
and back propagation. The flow chart of its training is shown in 
Fig.~\ref{FIG:WorkflowCNNs}. The whole training process includes (a) Forward Propagation and (b) Back Propagation. In forward propagation, six network models VGG16, VGG19, InceptionV3, Xception, ResNet50 and DenseNet201 are loaded first, and the weights of these neural networks are initialized. After that, the training data is input, and the output results are obtained through the convolutional layer, the lower sampling layer and the full connection layer. Next, the error between output and target value is calculated. When the error meets the requirements, the weights and thresholds of each layer are saved. If not, the process of back propagation is entered, which intends to update the weight to reduce error. If the performance does not improve, the learning rate will be decayed. Then return to start the next iteration. The process for each CNN is summarized in Algorithm~\ref{alg1}.}
\begin{figure}[!h] 
	\centering 
	\includegraphics[width=0.7\textwidth]{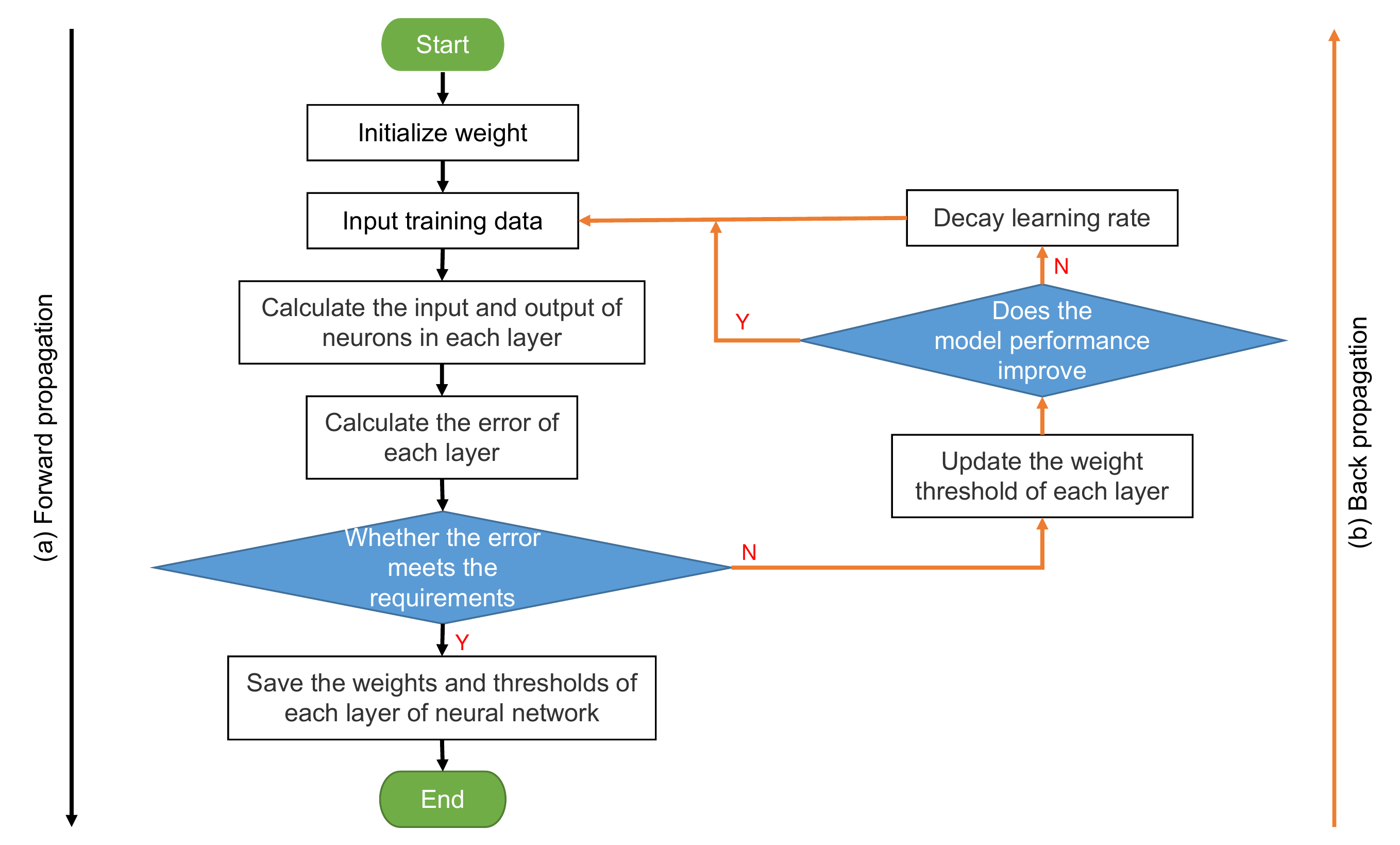}
	\caption{The workflow of neural networks.} 
	\label{FIG:WorkflowCNNs} 
\end{figure}

\begin{algorithm}[!h]  
\caption{Transfer learning algorithm in this paper} 
\label{alg1} 
\KwIn{The labeled dataset $S$, a base learning \textbf{Classifier} pre-trained on 
ImageNet, and the maximum number of interations $N$.} 
\textbf{Initialize} the initial weight vector, that $w_{jk}^{l}$ is the weight from the $k^{th}$ neuron in the $(l-1)^{th}$ layor to the $j^{th}$ neuron in the $l^{th}$ layer. $b^{l}_{j}$ and $a^{l}_{j}$ are the bias and the the activation (the output of the activation function) of the $j^{th}$ neuron in the $l^{th}$ layer, respectively.

\While{this model is improved}
{
    \If{not first time executing this code}
{
Attenuating the learning rate
}

    \For{$t=1, ..., N$}
{
Suppose the input is X and the output is Y. The value of $a^{l}_{j}$ depends on 
the activation of the previous layer of neurons: 
$a^{l}_{j}=\sigma(\sum_{k}w_{jk}^{l}a^{l-1}_{j}+b^{l}_{j})$\;
Rewrite the formular above into a matrix form: 
$a^{l}=\sigma(w^{l}a^{l-1}+b^{l})$\;
Call \textbf{Classifier}, calculate the activation using 
$a^{l}=\sigma(w^{l}a^{l-1}+b^{l})$ layer by layer, $X\rightarrow \hat{Y}$ \;
Calculate the loss function: $C=\frac{1}{2n}\sum_{x}\left \| y(x)-a^{L}(x)\right \|^{2}$. 
Where $n$ is the total number of training samples $x$, $y=y(x)$ is the ground truth, 
$L$ is the number of layers of the network, and $a^{L}(x)$ is the output vector of 
the network\;
Calculate the value of $w$ when the error is the smallest by taking the partial d
erivative of $w_{jk}^{l}$\;
\textbf{Update} the weights in the direction of reducing errors\;
}
}

\KwOut{the CNN with trained weights and thresholds for each layer.} 
\end{algorithm}

\par{Apart from transfer learning, six CNNs are also trained from scratch as a comparative 
experiment. Table~\ref{TABLE:CNNsParameter} summarizes the parameters of these models. 
Among them, both Top-1 Accuracy and Top-5 Accuracy refer to the accuracy of each neural 
network on the ImageNet validation set. Depth refers to the topological depth of each 
neural network, including the activation layer and batch normalization layer.} 
\begin{table}[!h]
\centering
\caption{Overview of CNN parameters.}
	\begin{tabular}{lrllrr }    
		\toprule   
		Model \quad\quad&   Size   &   \quad\quad Top-1 Accuracy  & Top-5 Accuracy & \quad Parameters &  \quad\quad Depth \\    
		\midrule   
		VGG16       \quad\quad& 528 MB   &\quad\quad 71.3\% & 90.1\% &\quad 138,357,544 &\quad\quad 23    \\
		VGG19       \quad\quad& 549 MB   &\quad\quad 71.3\% & 90.0\% &\quad 143,667,240 &\quad\quad 26    \\ 
		InceptionV3 \quad\quad& 92 MB    &\quad\quad 77.9\% & 93.7\% &\quad 23,851,784  &\quad\quad 159   \\
		Xception     \quad\quad& 88 MB    &\quad\quad 79.0\% & 94.5\% &\quad 22,910,480  &\quad\quad 126   \\
		ResNet50    \quad\quad& 98 MB    &\quad\quad 74.9\% & 92.1\% &\quad 25,636,712  &\quad\quad ---   \\
		DenseNet201 \quad\quad& 80 MB    &\quad\quad 77.3\% & 93.6\% &\quad 20,242,984  &\quad\quad 201   \\
		\bottomrule
	\end{tabular}
\label{TABLE:CNNsParameter}   
\end{table}

\subsubsection{Classification performance evaluation}

\par{In the classification of breast histopathological images, some hyperparameters 
are set. After many trials, these values are taken when the validation set obtains 
the best result. The exponential decayed learning rate is adopted regardless of 
training from scratch or training based on transfer learning. The decay steps are 
set to 5, the decay rate is 0.1, the initial learning rate is $1 \times 10^{-4}$, 
and the adaptive moment estimation (Adam) is selected as the optimizer. The other 
hyperparameter settings in the experiment are as follows:}

\par{Training the network from scratch: The initial input size is $224 \times 224 \times 3$, 
and the batch sizes of the VGG16, VGG19, InceptionV3, Xception, ResNet50, and DenseNet201 networks are set to 32, 16, 26, 8, 26 and 16, respectively. The epoch is specified as 60.}

\par{Transfer learning network: The initial input size is $460 \times 460 \times 3$, 
the batch size is set to 16, and the epoch is 60. In Table \ref{TABLE:ScratchandTransfer}, 
the classification results obtained by the above six CNNs based on these two 
training methods are summarized respectively.}
\begin{table}[!h]
\centering
\caption{The result and prediction time of CNN models on validation set (unit, $\%$).}
\begin{tabular}{llrrrrr}
\toprule
Model                         & Training Mode         & \multicolumn{1}{l}{Accuracy} & \multicolumn{1}{l}{Precision} & \multicolumn{1}{l}{Recall} & \multicolumn{1}{l}{F1-score} & \multicolumn{1}{l}{Time (unit, $s$)} \\ \midrule
\multirow{2}{*}{VGG16}       & Training from scratch & 89.50                                     & 89.65                                      & 89.32                                   & 89.48                                     & 17.28                                          \\
                              & Transfer Learning     & \textbf{95.49}                            & \textbf{95.40}                             & \textbf{95.58}                          & \textbf{95.49}                            & 32.50                                          \\
\multirow{2}{*}{VGG19}       & Training from scratch & 91.34                                     & 90.16                                      & 92.82                                   & 91.47                                     & 17.96                                          \\
                              & Transfer Learning     & \textbf{95.03}                            & \textbf{93.27}                             & \textbf{97.05}                          & \textbf{95.13}                            & 38.61                                          \\
\multirow{2}{*}{InceptionV3} & Training from scratch & \textbf{94.48}                            & 90.59                                      & \textbf{99.26}                          & \textbf{94.73}                            & 42.75                                          \\
                              & Transfer Learning     & 93.83                                     & \textbf{97.41}                             & 90.06                                   & 93.59                                     & 52.99                                          \\
\multirow{2}{*}{Xception}     & Training from scratch & 89.96                                     & 86.29                                      & 95.03                                   & 90.45                                     & 31.04                                          \\
                              & Transfer Learning     & \textbf{96.59}                            & \textbf{94.54}                             & \textbf{98.90}                          & \textbf{96.67}                            & 39.45                                          \\
\multirow{2}{*}{ResNet50}    & Training from scratch & 84.71                                     & 94.15                                      & 74.03                                   & 82.89                                     & 34.97                                          \\
                              & Transfer Learning     & \textbf{98.90}                            & \textbf{98.72}                             & \textbf{99.08}                          & \textbf{98.90}                            & 42.72                                          \\
\multirow{2}{*}{DenseNet201} & Training from scratch & 88.03                                     & 95.19                                      & 80.11                                   & 87.00                                     & 94.14                                          \\
                              & Transfer Learning     & \textbf{98.25}                            & \textbf{96.95}                             & \textbf{99.63}                          & \textbf{98.27}                            & 103.06                                         \\ \bottomrule
\end{tabular}
\label{TABLE:ScratchandTransfer}   
\end{table}

\par{From the statistical table of classification results, it can be concluded that 
the effect of transfer learning is generally better than that of training from scratch. 
Through transfer learning, the accuracy can be increased by around $3\%$ to $14\%$, 
which is a very satisfactory result. In addition, it is also found that training the 
InceptionV3 network from scratch is better than any other network using this method 
for training. At the same time, the accuracy of InceptionV3 training from scratch 
is $0.65\%$ higher than that of transfer learning. Both the recall and the F1-score 
are also higher than using transfer learning, with the values of $99.26\%$ and 
$94.73\%$, respectively. Nevertheless, this does not affect our subsequent choice 
of InceptionV3 based on transfer learning for the next step. This is because the 
training time of the network from scratch is generally more extended, and the 
performance of the computer should also be taken into account. In the same environment, 
training each network from scratch usually takes more than 2 hours. Therefore, transfer 
learning is applied to all the networks for the following research.}

\subsubsection{Overall evaluation and visual analysis transfer learning networks}

\par{Fig.~\ref{FIG:TrainCurve} shows the transfer learning training curves of the 
six CNNs: VGG16, VGG19, InceptionV3, Xception, ResNet50 and DenseNet201, 
respectively, to observe the changes in the training process of these models. 
It includes training set accuracy curve (blue curve), training set loss curve (green 
curve), validation set accuracy curve (yellow curve), validation set loss curve (red 
curve). As the number of training rounds increases, the accuracy curve shows an upward 
trend, and the loss curve shows downward. This is what we expected, which means that 
the performance of the models is gradually becoming stable.}
\begin{figure}[!htbp]
	\centering 
	\includegraphics[width=0.98\textwidth]{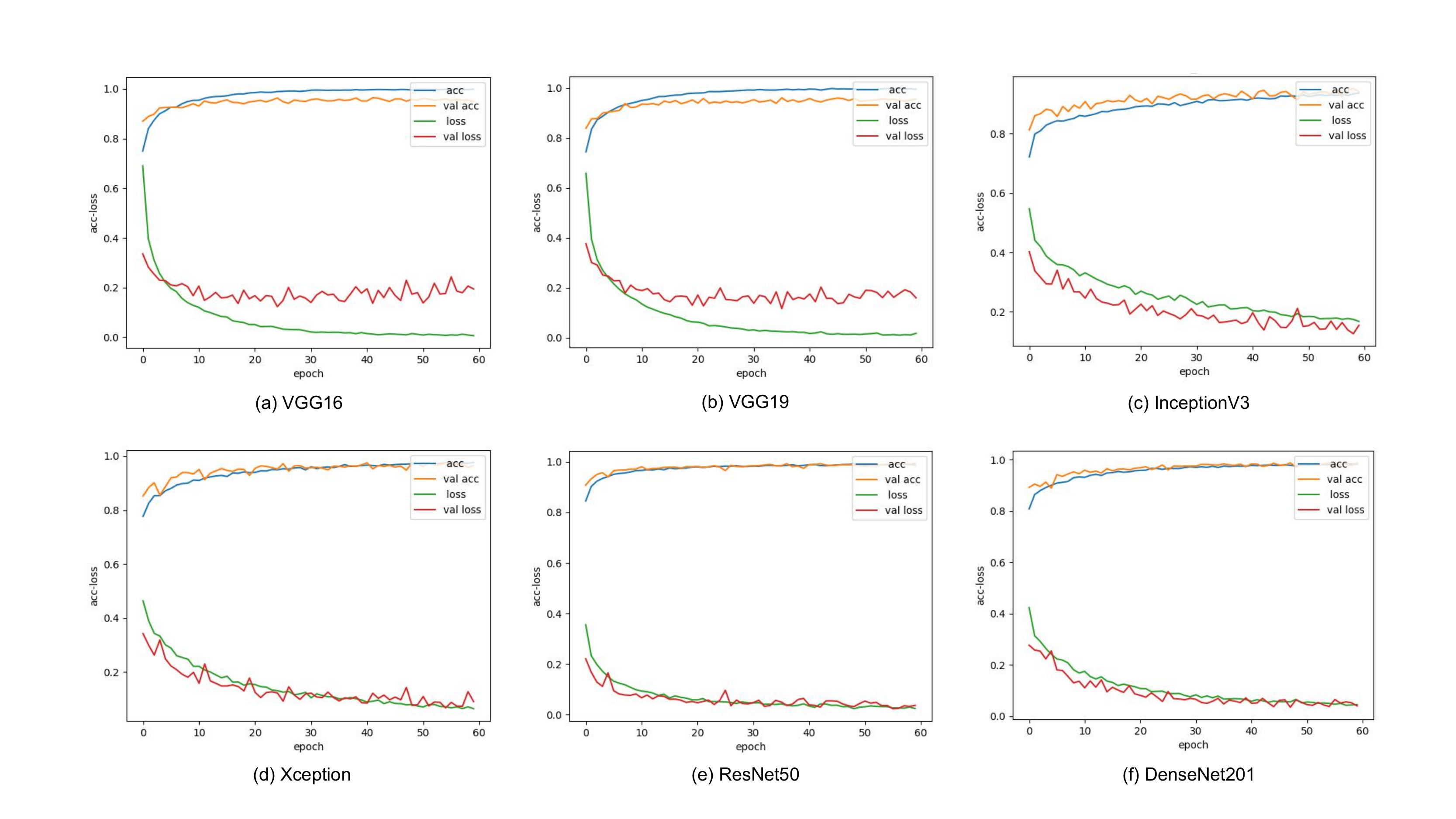}
	\caption{CNN training process curve.} 
	\label{FIG:TrainCurve}
\end{figure}

Subsequently, these six networks are compared, and their confusion matrices are 
shown in Fig.~\ref{FIG:CNNsMatrix}.
\begin{figure}[!h]
	\centering 
	\includegraphics[width=0.98\textwidth]{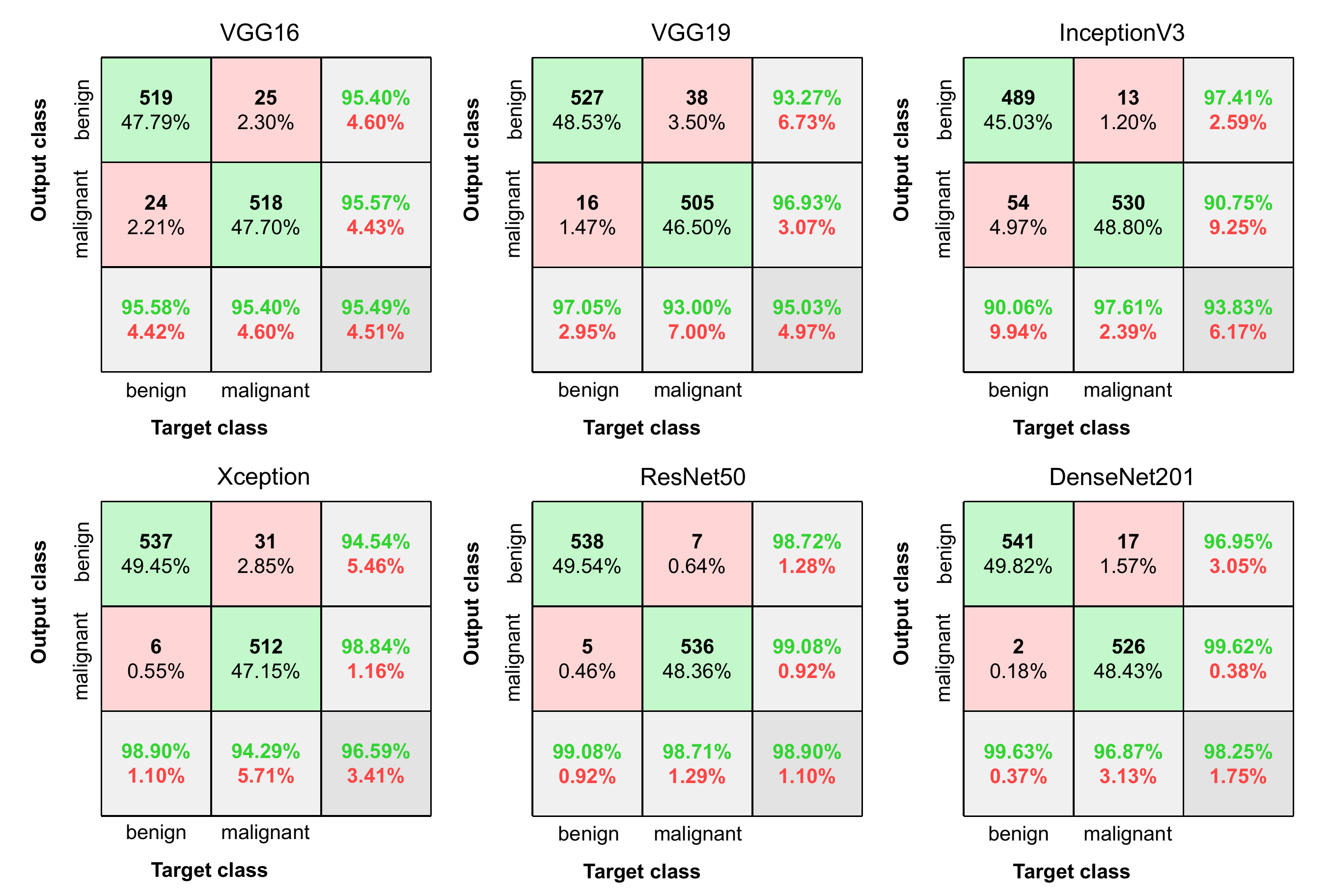}
	\caption{Confusion matrix of CNN on validation set.} 
	\label{FIG:CNNsMatrix} 
\end{figure}

\par{In the end, the ResNet50 network performs best, with three indicators that top 
all network models. To be precise, the accuracy is $98.90\%$, the precision reaches 
$98.72\%$, and the F1-score reaches $98.90\%$. The second place is the DenseNet201 
network, which is comparable to the ResNet50 network, at around $98\%$. Moreover, 
the recall of the DenseNet201 network is the highest among all models, reaching 
$98.63\%$. In order to analyze the classification results more 
intuitively, their evaluation indicators are drawn into a histogram, as shown in Fig.~\ref{FIG:BarChart}. In addition, we integrate the metrics and further calculate the average precision (AP) to evaluate the overall results. AP originates from the field of information retrieval and is primarily used to evaluate ranked lists of retrieved samples. The definition of AP in the experiment is shown in the Eq.~\ref{Eq:AP}:

\begin{equation}
AP = \frac{\sum_{n}^{i=1}(P(t))\times rel(t)}{N}
\label{Eq:AP}
\end{equation}

\par{$N$ is the number of relevant histopathology images, $P(t)$ is the $t$-th position in the list divided by considering the cutoff position, and $rel(t)$ is an index. The image ranking of the $t$-th position is the target type image, then it takes 1; otherwise, it takes 0. AP represents the average value of the image accuracy of the target type at the current location. Furthermore, we apply mean AP (mAP) to summarize the AP for each class. It is calculated by taking the average value of AP as shown in the table~\ref{TABLE:AP}. Except for InceptionV3, the other five networks all perform well, all of which are above 80\% in mAP.}

\begin{figure}[!h] 
	\centering 
	\includegraphics[width=0.75\textwidth]{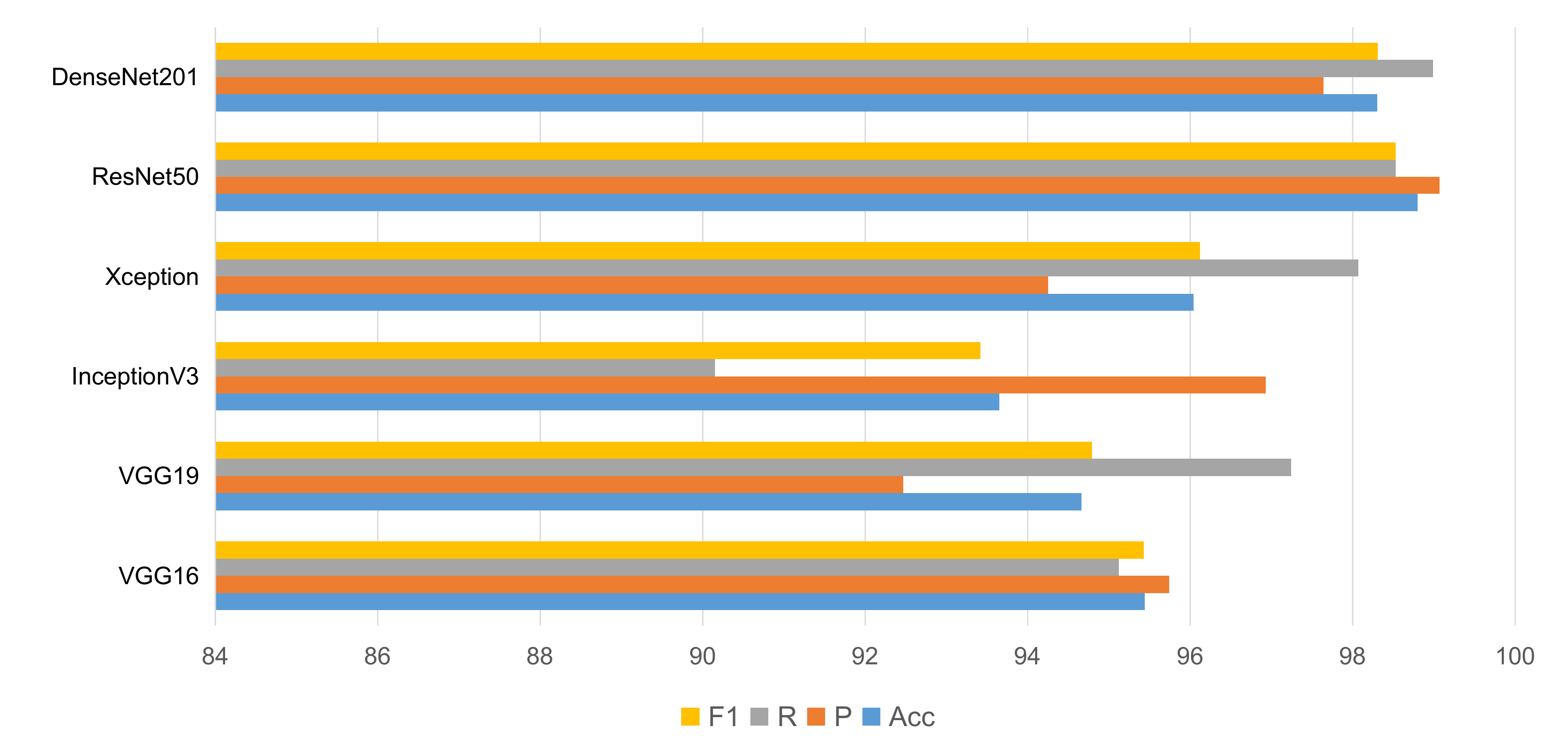}
	\caption{Comparison of evaluation indicators among six CNNs (unit, $\%$). The horizontal axis shows the value of the evaluation index, the vertical axis represents six different nodels, and the rectangular bars of different colors represent four evaluation indicators.}
	\label{FIG:BarChart} 
\end{figure}

\begin{table}[!h]
\centering
\caption{The AP and mAP of each model on validation set (unit, $\%$).}
\setlength{\tabcolsep}{7mm}{
\begin{tabular}{lrrr}
\toprule
Model       & AP (malignant) & AP (benign) & mAP     \\ \midrule
VGG16       & 83.79          & 79.30       & 81.55 \\
VGG19       & 86.52          & 84.54       & 85.53 \\
InceptionV3 & 70.56          & 71.10       & 70.83 \\
Xception    & 84.40          & 85.40       & 84.90 \\
ResNet50    & 85.00          & 84.47       & 84.73 \\
DenseNet201 & 83.27          & 86.10       & 84.68 \\ \bottomrule
\end{tabular}}
\label{TABLE:AP} 
\end{table}

\par{In the final evaluation, the results of the InceptionV3 network are relatively 
the worst, in which the accuracy, recall, F1-score and mAP are the lowest among the six 
models, with $94.48\%$, $90.06\%$, $93.59\%$ and 70.80\%, respectively. The lowest precision is obtained on VGG19, which is $93.27\%$. For networks that belong to 
the same series, such as the VGGNet series, the overall performance of the VGG16 network 
is better than that of the VGG19 network. Three indicators calculated in the VGG16 
network are about $0.3\%$ to $2\%$ higher than those in the VGG19 network. For the 
Inception series, the performance of the Xception network is better than that of the 
InceptionV3 network, and three of the evaluation indicators are about $2\%$ to $8\%$ 
higher than the InceptionV3. The performance of the DenseNet201 network and the 
ResNet50 network are relatively good, with satisfactory classification results. 
Finally, referring to the size, parameter amount, and network depth of the CNNs 
in Table~\ref{TABLE:CNNsParameter}, four networks are chosen in the end, namely 
VGG16, Xception, ResNet50 and DenseNet201 for the follow-up ensemble experiment.}

\subsection{Ensemble learning}

\subsubsection{Ensemble pruning}

\par{Ensemble pruning refers to selecting a subset of the individual learners that 
have been trained instead of all learners for the next step. It has two advantages: 
First, the ensemble result can be obtained using a smaller network scale. This can not 
only reduce the storage overhead brought by storing model, but also reduce the 
computing overhead corresponding to the output of the individual learner, thereby 
improving the efficiency of the model. In addition, the generalization performance 
of ensemble pruning is even better than the ensemble obtained by using all individual learners.}

\par{In this paper, two classifiers with relatively poor performance are pruned from 
the six CNN models based on transfer learning, and four classifiers are selected as 
individual learners, namely VGG16, Xception, ResNet50, and DenseNet201 
models. Since the accuracy is the main evaluation index of this experiment, the $Accuracy$ of each network is used as the weight of our voting method. Firstly, each weight is quantized with the step length 0.01 into \{0, 0.01, 0.02, ..., 0.99, 1\}. Then, in a weigh vector $w_{k=1,2,...,6}$, the combination of $w_{(k,1)}$, $w_{(k,2)}$, $w_{(k,3)}$ leads to the highest matching accuracy $Accuracy_{k}$ is chosen. Finally, the selected $w_{k=1,2,...,6}$ are implemented the ensemble learning method. And finally, we got the best ensemble accuracy of 98.90$\%$ in this way.}

\par{Thus, the final experimental procedure is obtained. First, in order to achieve 
data balance, data augmentation is applied to the BreaKHis dataset. The method used 
in this procedure is horizontal and vertical mirror flip. Then, the augmented data 
is input into each transfer learning based CNN, and six individual classifiers are 
obtained. After that, four classifiers with relatively good performance are selected and trained using the ensemble learning strategy of weighted 
voting. Finally, the ensemble results are evaluated.}

\subsubsection{Evaluation of ensemble learning algorithms}

\par{After training, the four indicators of accuracy, precision, recall, and F1-score 
are still used to evaluate the overall performance of the system. 
Fig.~\ref{FIG:EnsembleMatrix} shows the confusion matrix of ensemble learning. It 
can be found that the algorithm based on ensemble learning predicts 1 sample that 
should be benign as malignant and predicts 11 samples that should be malignant 
as benign.} 
\begin{figure}[!h] 
	\centering 
	\includegraphics[width=0.40\textwidth]{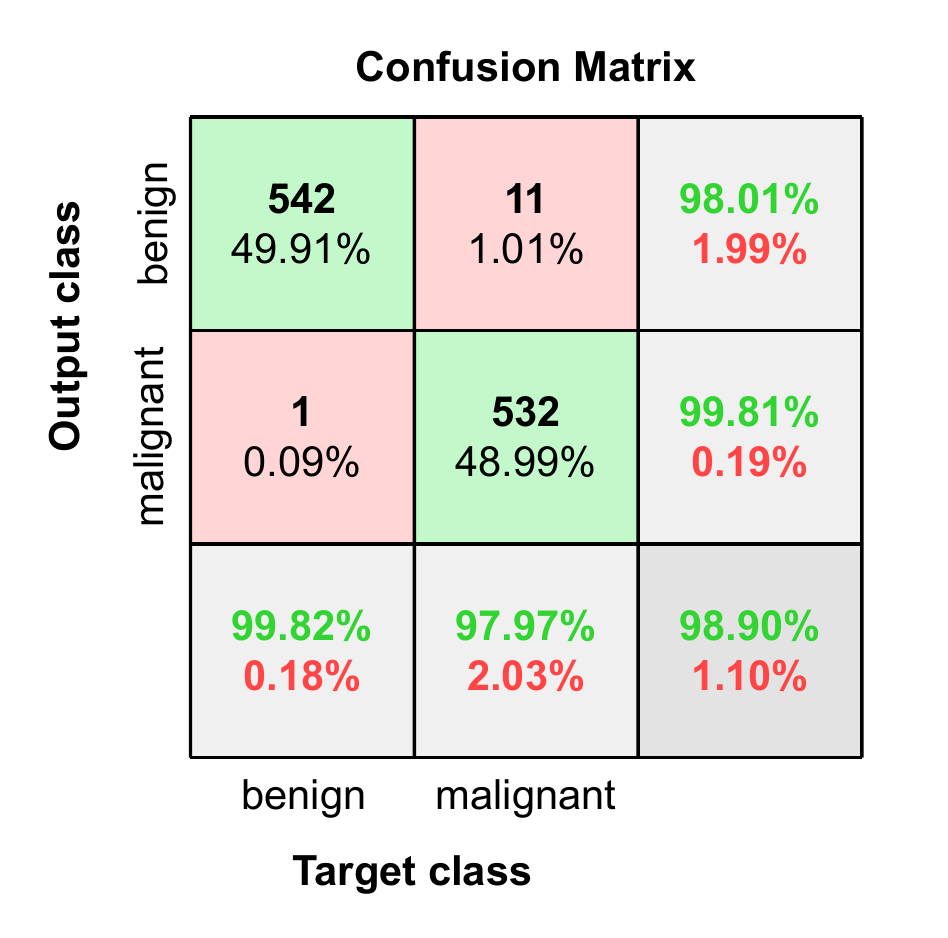}
	\caption{Confusion matrix of ensemble learning on the validation set.} 
	\label{FIG:EnsembleMatrix} 
\end{figure}

\par{Finally, the overall indicators of all single CNNs and our ensembled method 
are evaluated on the test set and  summarized in Table~\ref{TABLE:TestResult}. The 
ensemble learning strategy finally achieves $98.90\%$ accuracy, $98.72\%$ precision, 
$99.08\%$ recall, $98.90\%$ F1-score and 98.63\% mAP. Obviously, the accuracy, recall, F1-score and mAP
are higher than using any of the six CNNs based on transfer learning alone. After 
comparison, this method effectively improves the accuracy by $0.1\%$ to $5.25\%$, 
the recall by $0.09\%$ to $8.93\%$, the F1-score by around $0.3\%$ to $5\%$, and mAP by around 0.2\% to 20\%. In 
terms of precision, although it does not exceed the $99.07\%$ achieved by the 
ResNet50 network, its performance surpasses the other five models and wins second 
place. In summary, the classification performance of the ensemble learning method 
is generally better than that of the single transfer learning method.}
\begin{table}[!h]
\centering
\caption{Summary of classification results in testing process (unit, $\%$).}
\setlength{\tabcolsep}{7mm}{
	\begin{tabular}{llllll}    
		\toprule  
                        & \multicolumn{4}{l}{ Evaluation metrics}                                                                                                                        \\ 
\multirow{-2}{*}{Model} & Accuracy  & Precision       & Recall          & F1-score        & mAP       \\ \midrule  
VGG16             & 95.44           & 95.74           & 95.12           & 95.43           & 96.45          \\
VGG19             & 94.66           & 92.47           & 97.24           & 94.79           & 95.89        \\
InceptionV3       & 93.65           & 96.93           & 90.15           & 93.42           & 77.12        \\
Xception          & 96.04           & 94.25           & 98.07           & 96.12           & 96.88        \\
ResNet50          & 98.80           &{\textbf{99.07}} & 98.53           & 98.53           & 95.57        \\
DenseNet201       & 98.30           & 97.64           & 98.99           & 98.31           & 98.47        \\
Ensemble          &{\textbf{98.90}} & 98.72           &{\textbf{99.08}} &{\textbf{99.90}} & {\textbf{98.63}} \\ 
\bottomrule   
	\end{tabular}}
\label{TABLE:TestResult} 
\end{table}

\section{Discussion}
\label{section:4}

\subsection{Comparative experiment}

\par{To evaluate the ability of the breast histopathological image classiﬁcation algorithm proposed in this paper, a comparative experiment on Transformer and MLP models is carried out using the dataset divided by this experiment. The latest methods are summarized in Table.~\ref{TABLE:Comparison}, which have shown great power in natural image classification. However, their generalization ability has not been developed well like CNN for the specific histopathological images, and the overall classification performance is not so good. Especially Transformer, because of its good ability to describe global information, there should be improved versions suitable for pathological images in the near future, making novel contributions to this field.}

\begin{table}[!h]
\centering
\caption{Comparison of accuracy between the existing method and our method on the same BreaKHis dataset.}
\begin{tabular}{llrr}
\toprule 
\multicolumn{2}{l}{Model}                & Accuracy (unit, $\%$)    & Training time (unit, $s$) \\ \midrule 
\multirow{7}{*}{Transformer} & BoTNet-50 \cite{srinivas2021BTFVR}   & 90.75      & 4502     \\
                             & CaiT \cite{touvron2021going}         & 96.70      & 5081     \\
                             & CoaT \cite{xu2021CCIT}               & 92.91      & 420      \\
                             & DeiT \cite{touvron2021TDITD}         & 90.51      & 2101     \\
                             & LeViT \cite{graham2021levit}         & 93.42      & 13321    \\
                             & ViT \cite{dosovitskiy2020AIIWW}      & 80.11      & 1242     \\
                             & T2T-ViT \cite{yuan2021tokens}        & 80.02      & 2370     \\ \midrule 
\multirow{3}{*}{MLP}         & MLP-mixer \cite{tolstikhin2021mlp} & 83.84      & 5541     \\
                             & gMLP \cite{liu2021pay}              & 93.88      & 8407     \\
                             & ResMLP \cite{touvron2021RFNFI}       & 79.56      & 10341    \\ \midrule
Our method                   & Ensemble transfer learning model    & 98.90      & 3245       \\ \bottomrule
\end{tabular}
\label{TABLE:Comparison} 
\end{table}

\subsection{Comparison of previous research}

\par{As can be seen from Table.~\ref{TABLE:TestResult}, although compared to some single classifiers, the accuracy of our ensemble network has not been significantly improved. However, the network has been enhanced in the evaluation of the four indicators. In particular, F1-score is even close to 100$\%$, reflecting that both precision and recall of the model are excellent.}

\par{Fig.~\ref{FIG:Evaluation} shows the comparison of some images that are not correctly classified and those that are correctly classified. It can be found that the correctly classified image contains almost all classification information of benign and malignant so that the proposed algorithm can easily identify it. There are two main reasons why the image is mispredicted. First, this type of image has a very high degree of similarity to another type, such as Fig.~\ref{FIG:Evaluation} (a) and Fig.~\ref{FIG:Evaluation} (b). When one type of image is similar to another type of texture, distribution and color, it will not be easy to identify correctly. In addition, the wrongly predicted images often contain very little information to distinguish between benign and malignant, such as Fig.~\ref{FIG:Evaluation} (c) and Fig. \ref{FIG:Evaluation} (d). Patch-based images are allowed to contain many blanks during segmentation. The abovementioned reasons can interfere with the model's correct classification of benign and malignant breast histopathological images.} 
\begin{figure}[!htbp] 
	\centering 
	\includegraphics[width=0.95\textwidth]{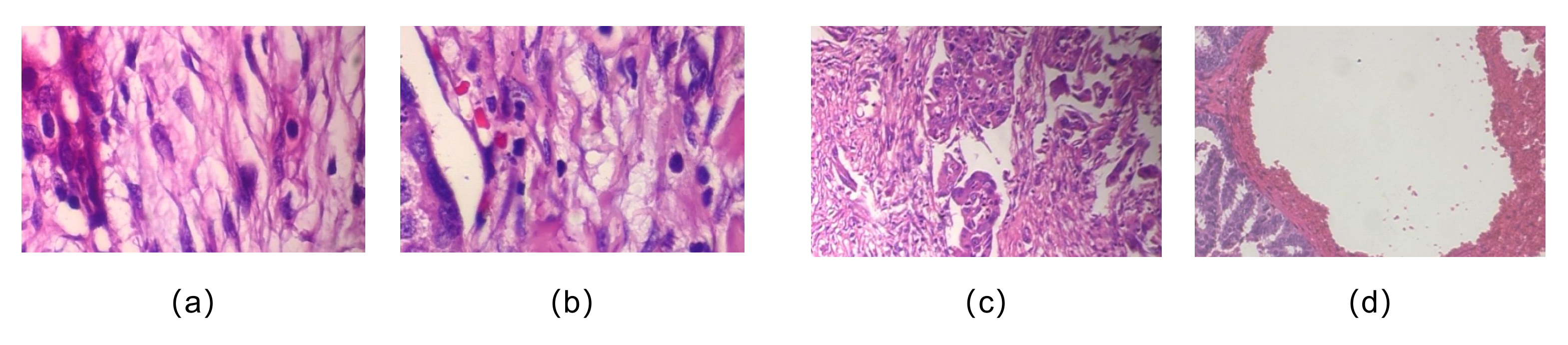}
	\caption{An example of the classification result. 
	(a) and (c) are correctly classified images. 
	(b) and (d) are wrongly classified images.} 
	\label{FIG:Evaluation} 
\end{figure}

\par{Additionally, we analyze the works based on deep learning on the BreaKHis dataset. Especially for image-level classification in previous research, such as~\cite{Li2020Imagelevel} with around $96\%$ and~\cite{hu2021Imagelevel} with $94\%$, our method with 98.90$\%$ accuracy works more effectively. Other ANNs and their results in different tasks are summarized in Table.~\ref{tbl8}, which are all applied in BreaKHis dataset. It should be noted that some studies may have done both binary and multiple classifications, but only the former results are considered. When compiling the tables, it is found that only a few studies used evaluation indicators other than accuracy. Even if the same indicator is employed (specifically, precision, recall and F1-score), the proposed method in this paper has a better performance. To facilitate statistics, only the accuracy of each study is discussed in detail. From the table, it can be concluded that in these algorithms, the accuracy of our method is effectively improved by 0.13$\%$ to 21.4$\%$.}

\begin{table}[!h]
\centering
\caption{Comparison of accuracy between the existing methods and our method based on BreaKHis dataset (unit, $\%$).}
\setlength{\tabcolsep}{7mm}{
\begin{tabular}{rllr}
\toprule
Year                     & Types of ANN that based on                        & Level   & Overall accuracy \\
 \midrule
2016                     & CNN \cite{Bayramoglu2016Deep}                     & patient & 83.25            \\
\multirow{2}{*}{2016}    & \multirow{2}{*}{AlexNet \cite{Spanhol2016Breast}} & patient & 84.53            \\
                         &                                                   & image   & 84.40            \\
\multirow{2}{*}{2017}    & \multirow{2}{*}{CaffeNet \cite{Spanhol2017Deep}}  & patient & 84.15            \\
                         &                                                   & image   & 83.80            \\
\multirow{2}{*}{2017}    & \multirow{2}{*}{CNN \cite{Song2017Adapting}}      & patient & 88.03            \\
                         &                                                   & image   & 85.33            \\
2017                     & CNN based on VGGNet \cite{Zhi2017Using}           & patient & 84.85            \\
2017                     & CNN \cite{Nejad2017Classification}                & patient & 77.50            \\
2017                     & CSDCNN \cite{Han2017Breast}                       & patient & 93.20            \\
\multirow{2}{*}{2017}    & \multirow{2}{*}{BiCNN \cite{wei2017deep}}         & patient & 97.41            \\
                         &                                                   & image   & 97.77            \\
2018                     & VGG16 \cite{Shallu2018Breast}                     & patient & 92.60            \\
2018                     & RNN and CNN \cite{Nahid2018Histopathological}     & patient & 91.00            \\
2018                     & DCNN \cite{Nahid2018Gistopathological2}           & patient & 92.19            \\
2018                     & CNN \cite{Du2018Breast}                           & image   & 90.00            \\
2018                     & DenseNet based CNN \cite{Nawaz2018Multi}          & patient & 95.40            \\
2018                     & ResNet \cite{Gandomkar2018Framework}              & patient & 98.77            \\
2018                     & DCNN \cite{Cascianelli2018Dimensionality}         & patient & 85.30            \\
2019                     & SA-Net \cite{Xu2019Look}                          & patient & 96.00            \\
2019                     & CNN \cite{Bhuiyan2019Transfer}                    & patient & 96.24            \\
2019                     & DCNN \cite{Xie2019Deep}                           & patient & 97.90            \\
2019                     & VGG16 and VGG19 \cite{Thuy2019Fusing}             & patient & 98.10            \\
\multirow{2}{*}{2020}    & \multirow{2}{*}{ResHist \cite{gour2020residual}}  & patient & 88.98            \\
                         &                                                   & image   & 88.02            \\
\multirow{2}{*}{2020}    & \multirow{2}{*}{DSoPN \cite{Li2020Imagelevel}}    & patient & 96.51            \\
                         &                                                   & image   & 96.54            \\
\multicolumn{1}{l}{2021} & myResNet-34 \cite{hu2021Imagelevel}               & image   & 94.03            \\
\multicolumn{1}{l}{2022} & Our method                                        & image   & 98.90            \\ 
\bottomrule
\end{tabular}}
\label{tbl8}   
\end{table}

\section{Conclusion and future work}
\label{section:5}

\par{In this paper, a framework that combines transfer learning and ensemble learning 
is proposed for the image-level classification of breast histopathological images. Finally, an accuracy of $98.90\%$ is obtained. First of all, based on the pre-segmented 
BreaKHis dataset, training from scratch and transfer learning are applied separately 
to train the six selected neural networks. Meanwhile, confusion matrices are used for 
the comparison and analysis of algorithms. Considering the two perspectives of less 
training time and high accuracy, we select the transfer learning method. After that, 
based on the idea of ensemble pruning, four networks are selected as individual 
classifiers. Through a large number of experiments, the weighted voting method with 
accuracy as the weight is used to combine these classifiers. Finally, we use the ten 
latest Transformer and MLP models to classify breast histopathological images on the 
same image-level dataset. It turns out that our method is quite competitive and ranks 
first in accuracy.}

\par{This research is dedicated to developing an algorithm to assist doctors in diagnosing 
tumor types correctly. The goal is to promote the calculation speed and efficiency, as 
well as the reliability of the final classification results. However, there is still 
massive potential for further exploration and development. Firstly, classifiers with better 
performance can be selected to further improve the final classification effect. Secondly, 
in addition to the weighted voting method used in this paper, more ensemble strategies 
and weights can be discussed in more depth. In addition, there are many types of breast 
cancer and many subtypes of tumors. This work is a two-class classification at 
the image level. In the future, a multi-classification system can be introduced on this 
basis. Henceforth, we are committed to continuously developing more efficient and 
high-accuracy classification models and will get involved in the image-level 
multi-classification tasks.}

\section*{Author Contributions}

Yuchao Zheng: method, experiment, result analysis and paper writing.
Chen Li: method, result analysis, paper writing and proofreading.
Xiaomin Zhou: method, experiment.
Haoyuan Chen: comparative experiment.
Hao Xu, Xinyu Huang and Marcin: data analysis, proofreading.
Yixin Li: comparative experiment.
Haiqing Zhang: paper writing.
Xiaoyan Li and Hongzan Sun: medical knowledge.
All authors contributed to the article and approved the submitted version.

\section*{Acknowledgements}
This work is supported by National Natural Science Foundation of China (No. 61806047). 
We thank Miss Zixian Li and Mr. Guoxian Li for their important discussion.

\section*{Conflict of Interest}
The authors declare that they have no conflict of interest in this paper.


\bibliographystyle{gbt7714-numerical}
\bibliography{refs}

\end{document}